\documentclass[12pt]{article}

\usepackage{scicite}

\usepackage{times}
\usepackage[hyphens]{url}  
\usepackage{graphicx}
\usepackage{amsmath}
\usepackage{amssymb}
\usepackage{booktabs}
\usepackage{array}
\usepackage{xcolor}
\definecolor{fmknowledge}{HTML}{0072B2}
\definecolor{fmdynamics}{HTML}{D55E00}
\definecolor{fmweak}{HTML}{009E73}
\graphicspath{{figures/}}

\newenvironment{sciabstract}{%
\begin{quote} \bf}
{\end{quote}}

\newcounter{lastnote}

\title{A small cognitive core explains human behavior at scale}

\author
{Marcel Binz$^{1,2\ast}$, Eric Schulz$^{1}$\\
\\
\normalsize{$^{1}$Institute for Human-Centered AI, Helmholtz Munich, Neuherberg, Germany}\\
\normalsize{$^{2}$Max Planck Institute for Biological Cybernetics, T\"ubingen, Germany}\\
\\
\normalsize{$^\ast$To whom correspondence should be addressed; E-mail: marcel.binz@tuebingen.mpg.de.}
}

\date{}

\begin{document}

\baselineskip24pt

\maketitle

\begin{sciabstract}
Psychology has proposed many principles of cognition. Each, however, was developed in isolation, leaving their combined explanatory power unknown.
We assembled a small set of established principles into one model, which we call COntextual Retrieval and Encoding (\texttt{CORE}).
\texttt{CORE} consists of a hierarchy of episodic memories into which prediction errors are written and whose encoding, retrieval, and forgetting depend on context.
Fitted jointly to 7,036,442 choices from 137 experiments, the model predicts human choices better than black-box models and reproduces 75\% of reported behavioral effects when simulated.
Moreover, its latent variables show classic signatures of Bayesian learning and event segmentation.
Taken together, these findings suggest that a small cognitive core accounts for much of human behavior.
They also open a path toward making psychology more than the sum of its parts, by building computational models on the entire experimental record.
\end{sciabstract}


\section*{Introduction}

A chief aim of cognitive science is to discover the computational principles that give rise to human behavior.
More than a century of research has produced many candidates, and there is much we already know about individual mechanisms that seem important.
For example, we know that associations strengthen when events co-occur \cite{hebb-organization-of-behavior-1949},
that learning is driven by prediction errors \cite{rescorla1972theory},
and that representations are organized hierarchically \cite{rumelhart1986parallel}.
What is missing, however, is an estimate of their combined explanatory power:
how much of human behavior can these principles, taken together, explain?

The most direct way to answer this question is to build a single computational model based on a small set of such principles
and test it against human behavior across a broad range of experiments \cite{newell1994unified, anderson2004integrated}.
The ingredients for this are now finally in place: large-scale data sets of human behavior,
expressive and meaningful computational models, and the compute to actually train them.
A model of this kind must meet two requirements. It must be built from cognitively meaningful principles, and it must capture behavior across paradigms.
To date, no model meets this standard. Existing approaches fall into
two categories: domain-specific cognitive models that are interpretable but
confined to a single paradigm \cite{rescorla1972theory, nosofsky2011generalized, ratcliff1978theory} and large-scale predictive models that capture behavior across paradigms but
remain black-box engines \cite{binz2025foundation}.

We introduce the \textbf{CO}ntextual \textbf{R}etrieval and \textbf{E}ncoding (\texttt{CORE}) model, combining the interpretability of cognitive models with the breadth of predictive models.
\texttt{CORE} consists of a hierarchy of episodic memories into which prediction errors are written
and whose encoding, retrieval, and forgetting are all context dependent.
Each individual component has a long history in cognitive science \cite{hebb-organization-of-behavior-1949, rescorla1972theory, rumelhart1986parallel, anderson1972simple, kohonen1972correlation, murdock1982theory, mcclelland1986trace},
but their joint explanatory power has not previously been tested across a broad range of human behavior.
\texttt{CORE} operates directly on natural language transcripts of experimental sessions,
allowing the same model to be fitted across paradigms.

We show that this small set of principles actually brings us quite far in explaining human behavior.
Fitted jointly to 137 experiments involving 7,036,442 choices
in 80,452 experimental sessions, \texttt{CORE} predicts human choices well compared to
standard machine learning architectures, such as Transformers \cite{vaswani2017attention},
and black-box foundation models of cognition, such as Centaur \cite{binz2025foundation}.
Furthermore, we find that, when simulated, \texttt{CORE} reproduces 75\% (50 out of 67) of behavioral effects identified in the corresponding studies.

Taken together, our work suggests that a surprisingly small cognitive core can account for much of human behavior,
and establishes a framework for testing theories of cognition not one experiment at a time,
but against large portions of the behavioral record at once.
We believe that this approach can move psychology from a loose collection of theories
toward an actually unified account of the human mind.

\section*{Results}

\paragraph*{COntextual Retrieval and Encoding (\texttt{CORE}).}
Let us suppose we wanted to build a cognitively meaningful model that captures human behavior across the full range of psychological experiments. Which components should such a model contain?

First, it needs to handle experiments that differ widely in their stimuli, instructions, and response formats.
Natural language provides an input format that allows us to express many different experimental paradigms, making it a suitable choice for the model's representation.
Next, the model needs a memory that can store experiences and retrieve them when they become relevant.
Importantly, what is being stored, retrieved, and forgotten should depend on the current context, because what is relevant may change over time.
The model should also rely on what it already knows instead of repeatedly storing the same information.
Finally, it should represent information in an efficient, hierarchical manner, from simple low-level features to more abstract concepts.

\begin{figure}[t]
\centering
\includegraphics[width=\textwidth]{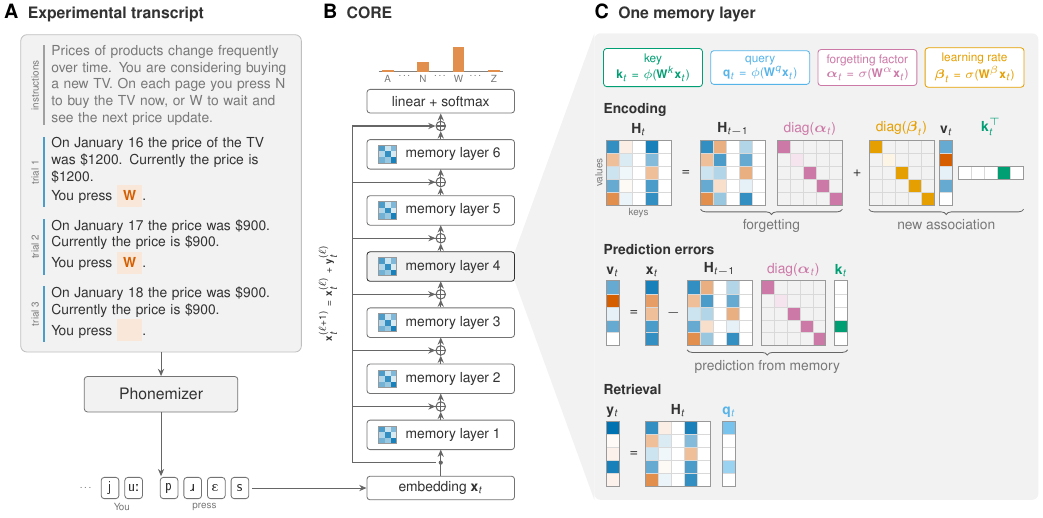}

\vspace{0.5em}
\begin{minipage}{\textwidth}
\refstepcounter{figure}\label{fig:model}
\noindent {\bf Fig.~\thefigure. The \texttt{CORE} model.}
(\textbf{A}) \texttt{CORE} reads the transcript of an experimental session (instructions, stimuli, feedback,
and the participant's own responses) converted into a sequence of phonemes.
(\textbf{B}) The phonemes are embedded and passed through a stack of memory layers, each of which adds its output to its input;
a linear read-out followed by a softmax predicts the next button press.
(\textbf{C}) Each memory layer computes a key, a query, a forgetting factor, and a learning rate from its current input, which makes every operation contextual.
Encoding: the memory is decayed feature-wise and a new association between a value and a key is added, scaled by the learning rate (Eq.~1).
Prediction errors: the value is the difference between the input and what the memory predicts for the current key (Eq.~4).
Retrieval: the query reads out the written values whose keys match it (Eq.~3).
\end{minipage}
\end{figure}

\texttt{CORE} brings these components together in a single model, building on ideas with a long history in cognitive science (Fig.~\ref{fig:model}).
It reads a transcript of an experimental session that contains all the information the participant had access to
(instructions, stimuli, previous responses, and feedback) and responds with button presses (A--Z). When processing an experiment, \texttt{CORE} first converts its transcript into a sequence of phonemes,
in line with models of speech perception that operate on phoneme-based representations \cite{mcclelland1986trace}.
Each phoneme $s_t$ is then mapped onto a learned $d$-dimensional embedding $\mathbf{x}_{t}$ \cite{bernard2021phonemizer, xu2021simple}. 

Following distributed accounts of human memory, information is written into an episodic memory that takes the form of an associative matrix $\mathbf{H}_{t} \in \mathbb{R}^{d \times d}$ \cite{anderson1972simple, kohonen1972correlation, murdock1982theory}.
In each time-step, the existing memory content $\mathbf{H}_{t-1}$ is multiplied by a feature-wise forgetting
factor $\boldsymbol{\alpha}_{t}$ and a new association between a value $\mathbf{v}_{t}$ and a key $\mathbf{k}_{t}$ is encoded using a Hebbian outer-product rule \cite{hebb-organization-of-behavior-1949, anderson1972simple, kohonen1972correlation, hopfield1982neural}:
\begin{equation}
\mathbf{H}_{t} = \mathbf{H}_{t-1}\,\mathrm{diag}\left(\boldsymbol{\alpha}_{t}\right) + \mathrm{diag}\left(\boldsymbol{\beta}_{t}\right)\mathbf{v}_{t}\,\mathbf{k}_{t}^{\top}
\end{equation}
What is written, and how strongly, depends on context \cite{howard2002distributed, polyn2009context}. Keys, forgetting factors, and learning rates $\boldsymbol{\beta}_{t}$ are all computed from the current input:
\begin{equation}
\begin{aligned}
\mathbf{k}_{t} &= \phi\left(\mathbf{W}^{k}\,\mathbf{x}_{t}\right) \\
\boldsymbol{\alpha}_{t} &= \sigma\left(\mathbf{W}^{\alpha}\,\mathbf{x}_{t}\right) \\
\boldsymbol{\beta}_{t} &= \sigma\left(\mathbf{W}^{\beta}\,\mathbf{x}_{t}\right)
\end{aligned}
\end{equation}
where $\phi$ denotes a nonlinearity
followed by normalization to unit length, and $\sigma$ a nonlinearity that maps onto values between zero and one.

To read from memory, \texttt{CORE} computes a context-dependent query $\mathbf{q}_{t} = \phi\left(\mathbf{W}^{q}\,\mathbf{x}_{t}\right)$.
Echoing classic models of memory retrieval \cite{hintzman1986schema}, it then returns the stored values, weighted by the similarity between their keys and the query:
\begin{equation}
\begin{aligned}
\mathbf{y}_{t} = \mathbf{H}_{t}\,\mathbf{q}_{t}
&= \Bigl(\,\sum_{i=1}^{t} \mathrm{diag}\left(\boldsymbol{\beta}_{i}\right)\mathbf{v}_{i}\,\bar{\mathbf{k}}_{i}^{\top}\Bigr)\,\mathbf{q}_{t}
= \sum_{i=1}^{t} \mathrm{diag}\left(\boldsymbol{\beta}_{i}\right)\mathbf{v}_{i}\,\Bigl(\bar{\mathbf{k}}_{i}^{\top}\,\mathbf{q}_{t}\Bigr) \\
\bar{\mathbf{k}}_{i} &= \mathrm{diag}\left(\boldsymbol{\alpha}_{t}\right) \cdots\, \mathrm{diag}\left(\boldsymbol{\alpha}_{i+1}\right) \mathbf{k}_{i}
\end{aligned}
\end{equation}
where $\bar{\mathbf{k}}_{i}$ is the key written at step $i$, with each feature decayed by the forgetting that has happened since.

What is written into memory are not the inputs themselves but --- as proposed in associative learning theory \cite{rescorla1972theory} ---
prediction errors between the input and what the memory retrieves for the current key \cite{schlag2021linear, yang2025gated}:
\begin{equation}
\begin{aligned}
\mathbf{v}_{t} &= \mathbf{x}_{t}
- \underbrace{\mathbf{H}_{t-1}\,\mathrm{diag}\left(\boldsymbol{\alpha}_{t}\right)\mathbf{k}_{t}}_{\text{prediction from memory}}
\end{aligned}
\end{equation}

Lastly, \texttt{CORE} stacks multiple of these associative memories on top of each other, reflecting the idea that cognition operates over multiple levels of representation \cite{rumelhart1986parallel}.
The output of the final layer is transformed with a linear mapping followed by a softmax into a probability for each button press.

We fit \texttt{CORE} using maximum a posteriori estimation
on \texttt{Psych-201}, a large-scale collection of psychological experiments from domains such as reinforcement learning,
decision-making, category learning, and social interaction \cite{binz2026post}.
For our analyses, we focus on experiments in which participants respond via button presses (137 experiments from 62 studies;
the training set comprises 7,036,442 choices in 80,452 experimental sessions).

To evaluate the resulting models, we rely on two metrics. The negative log marginal likelihood (NLML)
measures how improbable the data are when averaging over all possible weights of a model, weighted by their prior probability \cite{mackay1992practical, immer2021scalable}.
The predictive negative log-likelihood (PNLL) measures how well the fitted model predicts human choices in held-out sessions.
We compare \texttt{CORE} against standard machine learning architectures --- such as Transformers \cite{vaswani2017attention} ---
and black-box foundation models of cognition --- such as Centaur \cite{binz2025foundation} ---
as well as against ablated versions of itself.

\paragraph*{CORE captures human choices across paradigms.}

\begin{figure}[p]
\centering
\includegraphics[width=\textwidth]{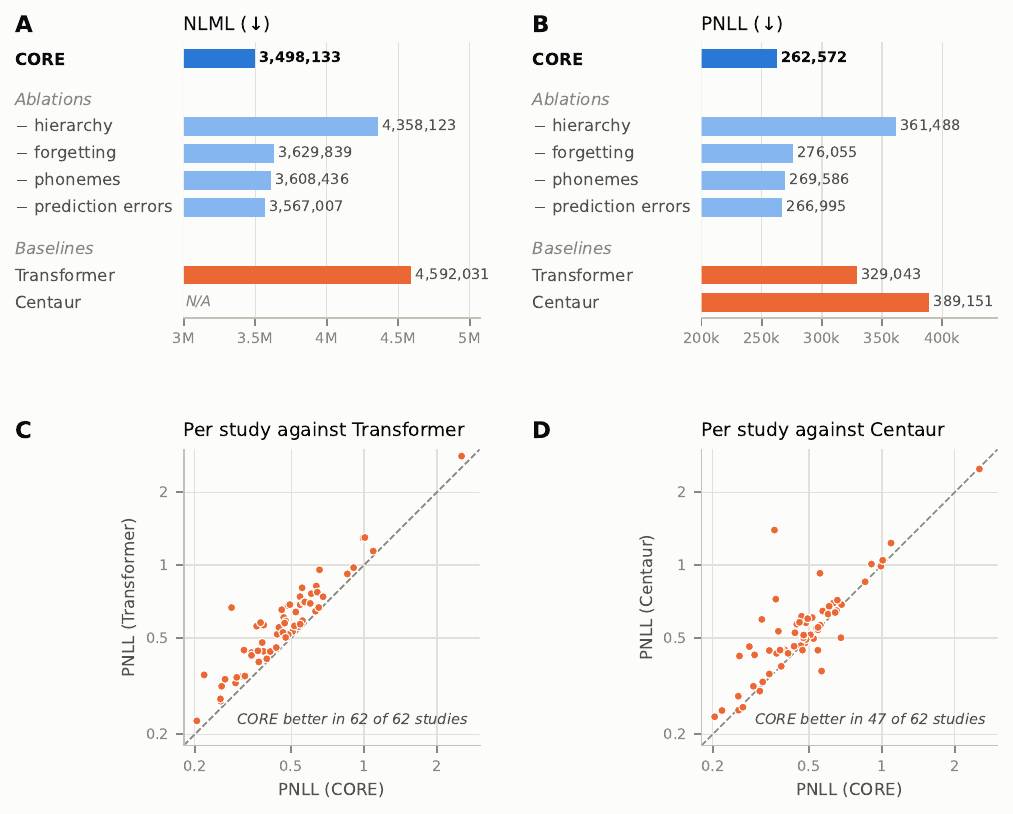}

\vspace{0.5em}
\begin{minipage}{\textwidth}
\refstepcounter{figure}\label{fig:comparison}
\noindent {\bf Fig.~\thefigure. Model comparison.}
(\textbf{A}) Negative log marginal likelihood (NLML) and (\textbf{B}) predictive negative log-likelihood (PNLL)
of \texttt{CORE}, its ablations, and the baselines; lower is better.
(\textbf{C}) PNLL per choice of \texttt{CORE} against the Transformer and (\textbf{D}) against Centaur in each of the 62 studies;
points above the diagonal are studies in which \texttt{CORE} predicts held-out choices better.
\end{minipage}
\end{figure}

We found that \texttt{CORE} predicts human choices well compared to several baselines (Fig.~\ref{fig:comparison}A--B).
It achieves both a lower negative log marginal likelihood (3,498,133 vs.\ 4,592,031) and a lower predictive negative log-likelihood (262,572 vs.\ 329,043)
than a Transformer-based sequence model of matched configuration.
\texttt{CORE} also predicts held-out choices better than Centaur (262,572 vs.\ 389,151) ---
a large language model fine-tuned on human behavior that has been shown to outperform domain-specific cognitive models.
These results hold broadly across studies: \texttt{CORE} predicts held-out choices better than the Transformer in all 62 studies, and better than Centaur in 47 of them (Fig.~\ref{fig:comparison}C--D).

\paragraph*{CORE relies on all of its components.}
To test whether each of \texttt{CORE}'s components is necessary, we removed them one at a time and compared \texttt{CORE} against each ablated version (Fig.~\ref{fig:comparison}A--B).
The hierarchy proved to be the most important component ($\Delta$ NLML $= 859{,}990$, $\Delta$ PNLL $= 98{,}916$).
That is because, with a single layer, the model lacks the expressivity to capture human behavior.
Forgetting came in second ($\Delta$ NLML $= 131{,}706$, $\Delta$ PNLL $= 13{,}483$).
In its absence, old associations never fade and can only be overwritten by new ones.
Phoneme-based encoding outperformed byte-pair encoding \cite{sennrich2016neural}, the standard tokenization scheme in modern language models ($\Delta$ NLML $= 110{,}303$, $\Delta$ PNLL $= 7{,}014$).
Even though the byte-pair model fit the training data better (negative log-likelihood of 2,904,051 vs.\ 3,066,566), it generalized worse and used considerably more embedding parameters (1,750,016 vs.\ 11,136).
Finally, writing prediction errors instead of raw inputs improved performance as well ($\Delta$ NLML $= 68{,}874$, $\Delta$ PNLL $= 4{,}423$).
Whereas writing raw inputs adds familiar content again, prediction errors contain only what the memory has not yet learned, thereby freeing up capacity for other content.

\paragraph*{CORE reproduces behavioral effects.}
Thus far, we have shown that \texttt{CORE} fits human choices well, which is a necessary but not sufficient requirement for a model of human cognition.
Equally important is that \texttt{CORE}, when simulated, reproduces the behavioral effects observed in humans \cite{palminteri2017importance}.
We therefore simulated \texttt{CORE} on the experiments of \texttt{Psych-201} and tested whether its simulated data show the same behavioral effects reported in the original studies.
To do this at scale, we used an agentic pipeline that implemented a simulator for each experiment and an analysis script that tests the paper's main effects on the resulting data.
In total, we obtained 67 effects from 47 experiments of 32 studies after filtering.
We simulated as many participants as in the original study and counted an effect as reproduced if it was statistically significant in the direction reported in the paper
(or, for predicted null effects, if it was not significant).

\begin{table}[p]
\centering
\begin{scriptsize}
\setlength{\tabcolsep}{4pt}
\begin{tabular}{>{\raggedright\arraybackslash}p{2.0cm} r l >{\raggedright\arraybackslash}p{6.1cm}}
\toprule
Domain & Reproduced & Experiment & Not reproduced \\
\midrule
Exploration      & 11/11 (100\%) & --- & --- \\
\addlinespace[8pt]
Reinforcement learning & 10/11 (91\%) & \texttt{chambon\_2020\_information}~\cite{chambon2020information} & \textcolor{fmweak}{Counterfactual feedback improves performance in a two-armed bandit.} \\
\addlinespace[8pt]
Risky and intertemporal choice & 9/13 (69\%)  & \texttt{olschewski\_2024\_frequent}~\cite{olschewski2024frequent} & \textcolor{fmknowledge}{Preference for the more frequent winner in a sampling task with equal-value options.} \\
\addlinespace[3pt]
                 &              & \texttt{olschewski\_2024\_frequent}~\cite{olschewski2024frequent} &\textcolor{fmknowledge}{Preference for the more frequent winner in a sampling task with identical outcome distributions.} \\
\addlinespace[3pt]
                 &              & \texttt{olschewski\_2024\_frequent}~\cite{olschewski2024frequent} &\textcolor{fmknowledge}{Preference for the right-skewed option in a sampling task without a frequent winner.} \\
\addlinespace[3pt]
                 &              & \texttt{spektor\_2024\_absolute}~\cite{spektor2024absolute} & \textcolor{fmweak}{Fewer risky choices in a loss-averse than in a gain-seeking context.} \\
\addlinespace[8pt]
Planning         & 4/6 (67\%)   & \texttt{kool\_2016\_when}~\cite{kool2016does} & \textcolor{fmdynamics}{Model-based planning signature in a two-step task.} \\
\addlinespace[3pt]
                 &              & \texttt{feherdasilva\_2023\_rethinking}~\cite{feher2023rethinking} & \textcolor{fmweak}{Intuitive task instructions increase model-based planning in a two-step task.} \\
\addlinespace[8pt]
Function and category learning & 9/14 (64\%)  & \texttt{flesch\_2018\_comparing}~\cite{flesch2018comparing} & \textcolor{fmdynamics}{Blocked training reduces interference between rules in context-dependent categorization (cardinal rules).} \\
\addlinespace[3pt]
                 &              & \texttt{flesch\_2018\_comparing}~\cite{flesch2018comparing} &\textcolor{fmdynamics}{Blocked training reduces interference between rules in context-dependent categorization (diagonal rules).} \\
\addlinespace[3pt]
                 &              & \texttt{flesch\_2018\_comparing}~\cite{flesch2018comparing} &\textcolor{fmdynamics}{Blocked training beats interleaved training in context-dependent categorization.} \\
\addlinespace[3pt]
                 &              & \texttt{binz\_2022\_heuristics}~\cite{binz2022heuristics} & \textcolor{fmdynamics}{No increase in accuracies over trials in multi-cue paired comparison with four unknown cues.} \\
\addlinespace[3pt]
                 &              & \texttt{binz\_2022\_heuristics}~\cite{binz2022heuristics} &\textcolor{fmweak}{Increased accuracies over trials in multi-cue paired comparison.} \\
\addlinespace[8pt]
Language and social judgment & 7/12 (58\%)  & \texttt{gnther\_2022\_patterns}~\cite{gunther2022patterns} & \textcolor{fmknowledge}{Existing compounds interpreted correctly in a relational judgment task.} \\
\addlinespace[3pt]
                 &              & \texttt{gnther\_2022\_patterns}~\cite{gunther2022patterns} &\textcolor{fmknowledge}{Novel compounds interpreted systematically in a relational judgment task.} \\
\addlinespace[3pt]
                 &              & \texttt{dentella\_2023\_systematic}~\cite{dentella2023systematic} & \textcolor{fmknowledge}{Ungrammatical sentences detected above chance in grammaticality judgment.} \\
\addlinespace[3pt]
                 &              & \texttt{cheung\_2024\_large}~\cite{cheung2025large} & \textcolor{fmweak}{Framing effect in sacrificial moral dilemmas.} \\
\addlinespace[3pt]
                 &              & \texttt{evangelidis\_2022\_upscaling}~\cite{evangelidis2023upscaling} & \textcolor{fmweak}{Inferior third option shifts preference toward quality over price in consumer choice.} \\
\midrule
Total            & 50/67 (75\%) & & \\
\bottomrule
\end{tabular}
\end{scriptsize}

\vspace{0.5em}
\begin{minipage}{\textwidth}
\refstepcounter{table}\label{tab:effects}
\noindent {\bf Table~\thetable.} Behavioral effects reproduced by simulating \texttt{CORE}, grouped by domain,
with every effect that was not reproduced listed. Missed effects are colored by failure mode:
\textcolor{fmknowledge}{requires knowledge from outside the experiment},
\textcolor{fmdynamics}{learning dynamics},
or \textcolor{fmweak}{right direction but not significant}.
\end{minipage}
\end{table}

\texttt{CORE} reproduced 50 of the 67 effects (75\%; Table~\ref{tab:effects}), indicating that it also mirrors qualitative patterns of human behavior.
We observed the best match in exploration and reinforcement learning,
likely because these paradigms are among the most prevalent in the training data.
\texttt{CORE} performed worst on language and social judgment.
We believe that this may be due to the fact that many of these tasks rely on knowledge that people acquire outside the laboratory,
about language, numbers, and the world.
\texttt{CORE} does not have access to such background knowledge because it has only ever seen transcripts of psychological experiments.
\texttt{CORE} also missed several effects that concern learning dynamics:
the model-based planning signature in one of three two-step experiments,
the benefits of blocked over interleaved training in context-dependent categorization,
and learning over trials in a condition where people did not learn.
Finally, six of the missed effects were in the right direction but did not reach significance.
We found that some of them are fragile in the human data as well: resampling the human participants at the original sample size
reproduced them in only 65--85\% of cases.

Looking at the ablated versions, we found that a variant without hierarchy reproduced only 22 of the 67 effects: it retained simple choice preferences, but lost most effects that require learning within an experiment.
This is expected, since a single layer can only recognize familiar tokens but not learn sequential contingencies between choices and outcomes.
The variant without forgetting reproduced 42 of the 67 effects.
It mainly lost effects that depend on the information of the current trial or context,
such as the dependence of risk taking on the current context, learning in contextual bandits, and the use of the current gamble's features in risky choice.

\paragraph*{CORE's latent variables are psychologically meaningful.}
\texttt{CORE}'s latent variables have a direct psychological interpretation.
To test whether they are indeed meaningful,
we recorded and analyzed them while the model performed the simulated experiments.

\begin{figure}[p]
\centering
\includegraphics[width=\textwidth]{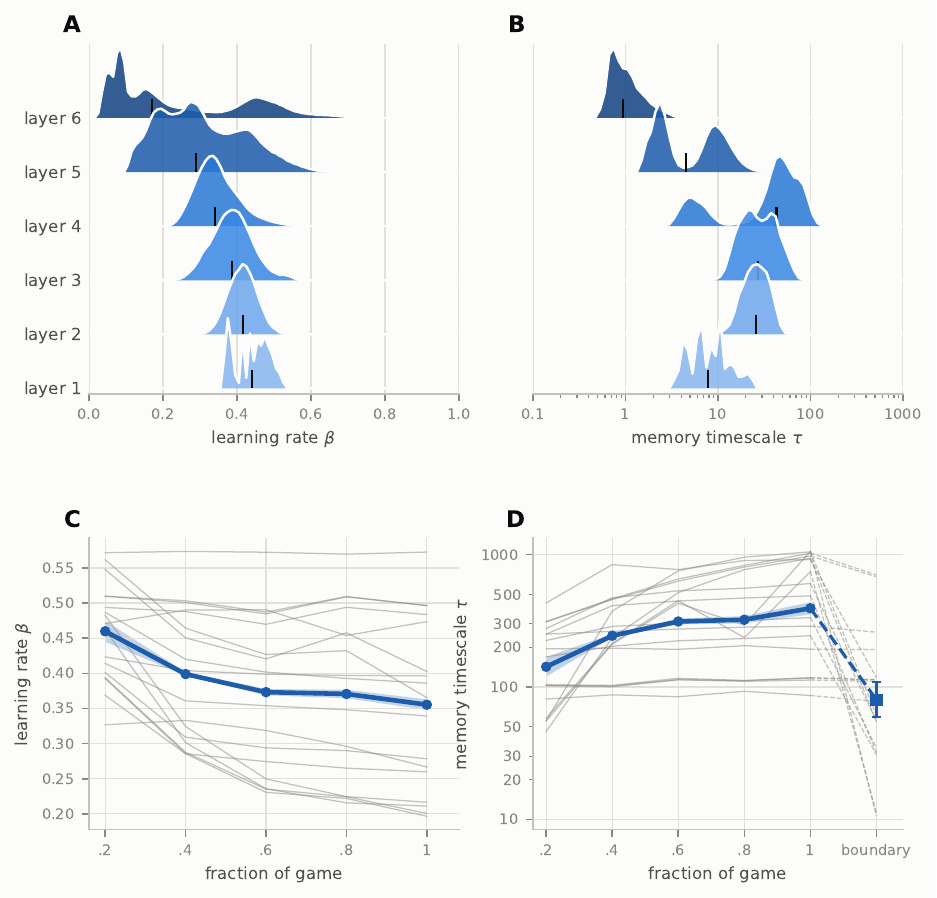}

\vspace{0.5em}
\begin{minipage}{\textwidth}
\refstepcounter{figure}\label{fig:gates}
\noindent {\bf Fig.~\thefigure. Latent variable analysis.}
(\textbf{A}) Learning rate $\boldsymbol{\beta}_t$ and (\textbf{B}) memory timescale $\tau = -1/\log \boldsymbol{\alpha}_t$ in tokens (log scale) per layer,
over all tokens of the simulated experiments; black ticks mark the medians.
(\textbf{C}) Learning rate and (\textbf{D}) memory timescale (log scale) of a single dimension of the fifth layer over the course of a game,
each the dimension with the largest consistent change (see Methods).
Values are averaged over the feedback that follows each choice.
Each game is divided into five bins of equal relative length.
The boundary is the text from the end of a game's last feedback up to the first choice of the next game.
Thick lines show the mean over experiments and thin lines show single experiments.
Bands and error bars show $\pm$ SEM over experiments of the change within participants.
\end{minipage}
\end{figure}

We found that both learning rates and memory timescales $\tau = -1/\log \boldsymbol{\alpha}_t$ (the number of tokens after which a stored association decays to $1/e$) varied within as well as across layers (Fig.~\ref{fig:gates}A--B).
Lower layers showed narrow, unimodal distributions, suggesting domain-general processes that are invariant across contexts.
Higher layers showed broad and often multimodal distributions, suggesting more specialized, context-specific processes.
Learning rates decreased with depth (Fig.~\ref{fig:gates}A; median 0.44 in the first vs.\ 0.17 in the last layer),
indicating that lower layers track the most recent input, whereas higher layers form running averages
that capture the structure of an experiment.
Memory timescales increased from the first to the fourth layer (Fig.~\ref{fig:gates}B; median 8 vs.\ 43 tokens).
Like for the learning rates, this points to a hierarchy from fast, local processing to slow, integrative processing.

Lastly, we asked whether \texttt{CORE}'s latent variables show signatures of classical learning and memory phenomena.
For this, we focused on experiments that consist of a sequence of independent games,
such as bandit problems with new reward probabilities or learning problems with a new underlying rule.
For every choice, we averaged the learning rate and memory timescale of the penultimate layer over the feedback that followed it.

Bayesian models of learning, such as the Kalman filter, predict that learning rates should decrease
as evidence accumulates \cite{kalman1960new, dayan2000explaining}.
\texttt{CORE}'s learning rates also exhibited this decline.
Fig.~\ref{fig:gates}C shows the dimension in which this effect was strongest.
For this dimension, learning rates decreased in 15 of the 17 experiments.
The decline was not limited to this dimension alone: learning rates decreased on average in 89 of the 128 dimensions of the layer.

Event segmentation theory holds that people maintain a model of the current event,
which they build up while the event unfolds and reset when a new event begins \cite{zacks2007event}.
Likewise, \texttt{CORE}'s memory timescales grew within a game and collapsed at the boundaries between games.
Fig.~\ref{fig:gates}D shows the dimension for which this effect was strongest:
its timescale dropped by 80\% at the boundary to the next game.
This effect was robust across experiments (17 of 17) and generalized across the layer:
on average, 94 of the 128 dimensions had a shorter timescale at the boundary than at the end of the game.

\section*{Discussion}

Psychology has no shortage of principles of human cognition, and new ones are added every year.
Many of them, however, are narrow in explanatory power, i.e., they fail to generalize beyond the specific paradigm they were developed for \cite{yarkoni2020generalizability}.
Instead of adding more principles, now is the time to ask a fundamentally different question: how far do the principles we already have take us in understanding the human mind?

To address this question, we took some of psychology's most established principles, assembled them into a single model,
and measured how much of human behavior they can actually explain.
We found that \texttt{CORE} predicts human choices better than standard machine learning architectures and a foundation model of human cognition with five orders of magnitude more parameters.
Even more strikingly, \texttt{CORE} reproduced 75\% of the behavioral effects reported in the original studies,
indicating that a small set of carefully selected principles can bring us quite far.

\texttt{CORE}'s main components --- phonemic input, associative memory, context dependence, prediction errors, and hierarchical representations ---
are all well established in cognitive modeling and neuroscience.
The concept of storing knowledge in associative memories has a long history, including the linear associator \cite{anderson1972simple, kohonen1972correlation}, Hebbian learning \cite{hebb-organization-of-behavior-1949}, Hopfield networks \cite{hopfield1982neural, krotov2016dense},
the context maintenance and retrieval model \cite{polyn2009context}, and others \cite{howard2002distributed}.
Prediction errors are central to many theories of human learning, from general frameworks such as predictive coding \cite{rao1999predictive}
down to specific implementations like the Rescorla--Wagner model \cite{rescorla1972theory}.
Hierarchical representations feature in many lines of research, including
the classic work on receptive fields and modern connectionist models \cite{rumelhart1986parallel}.
\texttt{CORE}'s components also map onto proposals for how the brain implements memory \cite{gershman2025key, irie2025fast}.
For instance, Kozachkov et al.\ have shown how associative memories --- like the one \texttt{CORE} uses --- could be implemented
by networks of neurons and astrocytes \cite{kozachkov2023building, kozachkov2025neuron}.
More broadly, \texttt{CORE} repeats a single type of memory layer across its hierarchy, much like the neocortex is thought to repeat a canonical local circuit across areas \cite{douglas2004neuronal, bastos2012canonical}.

\texttt{CORE} furthermore borrows ideas from machine learning, in particular linear attention \cite{katharopoulos2020transformers}, fast weight programmers \cite{schmidhuber1992learning, schlag2021linear}, and Gated Delta Networks \cite{yang2025gated, hatamizadeh2026gated}.
What distinguishes \texttt{CORE} from these models is what it leaves out: it contains no computation besides writing to and retrieving from memory,
which makes it possible to attribute its behavior to a handful of principles, and nothing else.
Parameter estimation at this scale relied on the flash-linear-attention library \cite{yang2024fla},
extended with a custom kernel that cut the run time from about 10 days to about 24 hours on a single GPU.

We have shown that a few established principles can take us quite far, but there is still some way to go. What is needed to capture the remaining 25\%?
\texttt{CORE} struggles with model-based planning, as shown by our model simulations \cite{daw2011model}.
This makes sense, given that it contains no explicit planning mechanism.
Future model iterations could incorporate such a mechanism in several ways, for example using ideas from looped Transformers \cite{dehghani2018universal}.
There are also mechanisms that our method cannot detect at present, simply because the data do not support it.
Long-term memory is one example. Experiments in our data set typically span less than an hour,
so how participants learn and store information over weeks and months is never observed.
Longitudinal data would allow us to add a long-term memory to \texttt{CORE},
which would make it possible to study how such a memory differs from \texttt{CORE}'s episodic memory and how consolidation transfers information from one to the other \cite{mcclelland1995there}.

Longitudinal data of this kind are easiest to collect outside the laboratory, in naturalistic environments that people revisit over weeks and months. While traditional cognitive models resist such environments,
\texttt{CORE} is stimulus-computable \cite{yamins2016using}: it operates directly on natural language
and could be readily extended to other sensory modalities.

\texttt{CORE}'s principles are interpretable, but its computations are not.
Instead of a full mechanistic account, \texttt{CORE} is, in its current form, best seen as an intermediate step between classical cognitive models and black-box systems \cite{peterson2021using}.
There are still components of \texttt{CORE} that lack a theoretical grounding, e.g., how it computes keys, values, learning rates, and forgetting.
Future models should retain the broad behavioral coverage of \texttt{CORE} while replacing learned transformations with independently motivated computational principles, thereby moving from an interpretable architecture toward a complete mechanistic account.

\section*{Conclusion}

Imagine a future in which every behavioral experiment ever conducted is available in a unified format.
How should we develop theories of the mind in such a world?
For most of its history, psychology has tested its theories one experiment at a time \cite{newell2012you}.
This leaves most of the evidence unused: each theory is constrained by a few experiments, not by the entire behavioral record of psychology.
What we have pursued amounts to a radically different paradigm: principles are no longer tested in isolation, but assembled into a single model that is fitted jointly to many experiments.

\texttt{CORE} is an initial step in this direction.
It shows that a small set of established cognitive mechanisms can account for a surprisingly broad range of human behavior when fitted jointly across experiments.
The opportunity, however, is larger than any single model.
If every experiment ever conducted constrains a shared model of the mind, psychology may finally become more than the sum of its experiments.

\section*{Materials and Methods}

\paragraph*{Data.}
We used a version of \texttt{Psych-201} re-transcribed by an agentic pipeline.
Each experiment was converted into a natural-language transcript using DeepSeek-V4-Flash (\texttt{deepseek-\allowbreak v4-\allowbreak flash-\allowbreak 0731}) \cite{xu2026deepseek}
and verified using Claude Fable~5.1 (\texttt{claude-\allowbreak fable-\allowbreak 5-\allowbreak 1}) \cite{anthropic2026fable51}.
We release this version on the Hugging Face Hub:
\url{https://huggingface.co/datasets/marcelbinz/Psych-201-discrete-agentic} (training set)
and \url{https://huggingface.co/datasets/marcelbinz/Psych-201-discrete-agentic-test} (test set).
Transcripts contain all the information participants had access to and were never truncated.
Human choices are enclosed in response markers (\texttt{[HUMAN\_\allowbreak RESPONSE]}\ldots\allowbreak\texttt{[/HUMAN\_\allowbreak RESPONSE]}).
We kept these markers only around button presses and removed them around all other responses,
which remain in the transcript as text but are not predicted; sessions without any button press were dropped.
The resulting data set comprises 137 experiments from 62 studies~\cite{akata2025playing,anvari2024testing,badham2017deficits,bahrami2020arm,barnby2022knowing,bavard2018reference,bavard2021two,binz2022heuristics,breslav2022shuffle,chambon2020information,chen2018age,cheung2025large,christian2026resolving,ciranka2025internal,cohen2020rational,cox2018information,decker2016creatures,dentella2023systematic,dezfouli2019models,dubois2022value,evangelidis2023upscaling,fan2023trait,feher2023rethinking,feng2021dynamics,flesch2018comparing,frey2017risk,garcia2023experiential,gershman2018deconstructing,gunther2020trying,gunther2022patterns,gunadi2022impact,haines2020anxiety,heffner2022probabilistic,jagadish2023zero,jansen2021rational,kool2016does,kool2017cost,krueger2024identifying,levering2020revisiting,moutoussis2018change,nussenbaum2023novelty,nussenbaum2024sensitivity,olschewski2024frequent,olschewski2025optimal,pirrone2024subjective,plonsky2025predicting,russek2024heuristics,sadeghiyeh2020temporal,schulz2020finding,shahar2019improving,singh2019training,spektor2024absolute,steingroever2015data,suthaharan2021paranoia,thoma2025emerging,vandendriessche2023contextual,van2021probabilistic,witte2024should,wu2023chunking,xu2021novelty,xu2023augmenting,zorowitz2023data}, with 7,634,315 choices in 83,352 sessions.
Each button press was represented by a single token, and the loss was computed on these tokens only.
To this end, we added one separate token per button (A--Z) to the vocabulary of the tokenizer,
so that a button press is never split into phonemes or subwords.
For each study, we held out $\min(100, \lfloor 0.1\,n \rfloor)$ of its $n$ sessions as a test set (597,873 choices in 2,900 sessions).

\paragraph*{Model.}
\texttt{CORE} converts each transcript into phonemes using the eSpeak-based phoneme tokenizer of \texttt{wav2vec 2.0} \cite{xu2021simple},
and maps each phoneme onto a learned 128-dimensional embedding.
The embeddings pass through six memory layers, each of which normalizes its input using RMSNorm \cite{zhang2019root} and adds the output of its memory to it.
Each layer consists of four heads, i.e., independent memory modules that operate in parallel.
The memory matrix of each head starts empty ($\mathbf{H}_0 = \mathbf{0}$) at the beginning of every session.
Each head computes its own keys, queries, learning rates, and forgetting factors (32 dimensions each) from the full 128-dimensional input;
the outputs of the four heads are concatenated to form the output of the layer.
Keys and queries are computed with a SiLU nonlinearity \cite{elfwing2018sigmoid} followed by normalization to unit length,
the learning rate $\boldsymbol{\beta}_t$ with a logistic function,
and the forgetting factor $\boldsymbol{\alpha}_t$ as $\exp\left(-a \cdot \mathrm{softplus}(z + b)\right)$ with learned $a > 0$ and $b$ per feature.
A linear read-out followed by a softmax maps the output of the last layer onto the response buttons.
The model contains no other components and has 452,608 parameters in total.
The memory update was implemented as a chunked Triton kernel extending the flash-linear-attention library \cite{yang2024fla}.

\paragraph*{Training.}
All parameters were fitted jointly on the training set by maximum a posteriori estimation,
using schedule-free AdamW \cite{defazio2024road}
with a learning rate of $10^{-3}$ and a decoupled weight decay of 0.5.
We trained for 100 epochs with batches of 304,500 tokens and a single random seed.
Training took about 24 hours on a single NVIDIA H100 GPU and used the \texttt{transformers} library \cite{wolf2019huggingface}.

\paragraph*{Model comparison.}
We approximated the log marginal likelihood with a Laplace approximation around the fitted weights,
using a diagonal empirical Fisher information matrix over all parameters and a prior precision per layer that was optimized after training \cite{daxberger2021laplace}.
The predictive log-likelihood was the summed log-probability of all held-out choices.
The Transformer baseline used the same configuration as \texttt{CORE} (128 dimensions, six layers, four heads)
but a byte-pair tokenizer \cite{sennrich2016neural}.
The ablations removed one component each: a single memory layer instead of six, no forgetting ($\boldsymbol{\alpha}_t = 1$),
Hebbian instead of error-driven encoding ($\mathbf{v}_t = \mathbf{x}_t$), and a byte-pair tokenizer instead of phonemes.
We did not compute the marginal likelihood for Centaur, because it is unclear how its pretraining should be accounted for.
Per-study likelihoods, behavioral effects, and example transcripts for all models can be explored at \url{https://core-cognition.github.io/core-model/results/}.

\paragraph*{Experiment simulations.}
For each experiment, an agentic pipeline produced a simulator of the task, which generates transcripts in the same format as the data,
and an analysis script that tests the main effects reported in the original paper.
We retained experiments for simulation and subsequent analysis based on the following criteria.
First, we excluded experiments whose simulator or original transcripts involve other responses,
such as numbers, words, or free text.
We then excluded effects whose analysis depended on information the simulators cannot produce,
such as reaction times, age, gender, ratings, or clinical group membership.
Finally, we kept only effects that the original human data reproduce under the same analysis.
In total, this yielded 67 effects from 47 experiments of 32 studies.
\texttt{CORE} was simulated by sampling its predicted distribution with a temperature of $1.0$.
We simulated as many participants as in the original study, up to 1,000.
Failure modes from Table~\ref{tab:effects} were assigned by hand after scoring.

\paragraph*{Latent-variable analysis.}
We analyzed the latent variables of \texttt{CORE} while it performed the simulated experiments.
For each simulated session, we recorded the learning rate $\boldsymbol{\beta}_t$ and the forgetting factor $\boldsymbol{\alpha}_t$
of every dimension and layer at every token.
Forgetting factors were converted into memory timescales $\tau = -1/\log \boldsymbol{\alpha}_t$ (i.e.,
the number of tokens after which a stored association has decayed to $1/e$ of its strength) and averages of memory timescales were computed on a logarithmic scale.

For Fig.~\ref{fig:gates}A--B, we averaged the learning rates and log forgetting factors over the dimensions of each head.
For each experiment, we computed histograms over all tokens and heads of a layer, and normalized them to sum to one.
We then averaged the histograms over the simulated experiments, so that each experiment has the same weight regardless of its number of participants or tokens.
Medians were computed from the averaged histograms.

To follow the latent variables over the course of a game, we used the simulated experiments that consist of a sequence of games
(at least four games per participant with at least five choices in a typical game).
This resulted in 17 experiments from 9 studies.
For Fig.~\ref{fig:gates}C--D, we kept the individual dimensions instead of averaging over the dimensions of a head.
For each choice, we averaged the latent variables over the feedback that follows it,
from the choice to the end of the line that reports the outcome.
For the boundary between two games, we averaged over the text between the feedback of a game's last choice and the first choice of the next game,
which introduces the new game and presents its first stimulus.
We assigned each choice to one of five bins according to its position in the game, relative to the length of the game,
so that the first and last choices of every game fall into the first and last bin;
games with fewer than five choices were excluded.
We averaged within participants, then over the participants of an experiment, and then over experiments,
so that each experiment has the same weight.
The shaded bands and error bars in Fig.~\ref{fig:gates}C--D show the standard error over experiments of the change within participants, that is, after we subtracted each participant's mean over the five bins.

\bibliography{refs}
\bibliographystyle{Science}

\section*{Acknowledgments}
{\bf Funding:} This work was supported by the Helmholtz Association's Initiative and Networking Fund on the HAICORE@FZJ partition.
E.S.\ is supported by the European Research Council (ERC) through the TACOS grant (grant agreement No.\ 101116962).
{\bf Author contributions:} M.B.\ conceived the study, developed the model, performed the analyses, and wrote the manuscript.
E.S.\ edited the final draft and acquired funding.
{\bf Competing interests:} The authors declare no competing interests.
{\bf Data and materials availability:} The \texttt{Psych-201} training and test sets are available on the Hugging Face Hub
at \url{https://huggingface.co/datasets/marcelbinz/Psych-201-discrete-agentic}
and \url{https://huggingface.co/datasets/marcelbinz/Psych-201-discrete-agentic-test}.
Code is available at \url{https://github.com/CORE-COGNITION/core-model}.

\end{document}